\documentclass[11pt,a4paper]{article}
\usepackage[hyperref]{emnlp2020}
\usepackage{times}
\usepackage{url}
\usepackage{latexsym}

\usepackage{xspace}
\usepackage{url}
\usepackage[shortlabels]{enumitem}
\usepackage{pifont}
\usepackage{amsmath}
\usepackage{textcomp}
\usepackage{float}
\usepackage{amssymb}
\usepackage[font=small,labelfont=bf]{caption}
\usepackage{xcolor,colortbl}
\usepackage{graphicx}
\usepackage{microtype}

\newcommand{\bert}{\textsc{bert}\xspace}
\newcommand{\bertbase}{\textsc{bert-base}\xspace}
\newcommand{\eurlexdata}{\textsc{eurlex57k}\xspace}

\newcommand{\nlp}{\textsc{nlp}\xspace}
\newcommand{\eurlex}{\textsc{eurlex}\xspace}
\newcommand{\mimiciii}{\textsc{mimic-iii}\xspace}
\newcommand{\icdix}{\textsc{icd-9}\xspace}
\newcommand{\eu}{\textsc{eu}\xspace}

\newcommand{\eurovoc}{\textsc{eurovoc}\xspace}

\newcommand{\scibert}{\textsc{scibert}\xspace}

\newcommand{\cls}{\texttt{[cls]}\xspace}
\newcommand{\sep}{\texttt{[sep]}\xspace}
\newcommand{\lmtc}{\textsc{lmtc}\xspace}

\newcommand{\lstm}{\textsc{lstm}\xspace}
\newcommand{\baseline}{\textsc{flat}\xspace}
\newcommand{\lastsix}{\textsc{last-six}\xspace}
\newcommand{\onebyone}{\textsc{one-by-one}\xspace}
\newcommand{\allinpairs}{\textsc{in-pairs}\xspace}
\newcommand{\skipboost}{\textsc{hybrid}\xspace}

\aclfinalcopy

\title{Layer-wise Guided Training for BERT:\\Learning Incrementally Refined Document Representations}

\author{Nikolaos Manginas$^{\;\dagger}$ \qquad Ilias Chalkidis$^{\;\dagger\;\ddagger}$ \qquad Prodromos Malakasiotis$^{\;\dagger\;\ddagger}$\\ $^{\dagger\;}$Institute of Informatics \& Telecommunications, NCSR ``Demokritos'' \\ $^{\ddagger\;}$Department of Informatics, Athens University of Economics and Business \\{\tt [nmanginas,ichalkidis,pmalakasiotis]@iit.demokritos.gr}}

\date{}

\begin{document}
\maketitle

\begin{abstract}
    Although \bert is widely used by the \nlp community, little is known about its inner workings. Several attempts have been made to shed light on certain aspects of \bert, often with contradicting conclusions. A much raised concern focuses on \bert's over-parameterization and under-utilization issues. To this end, we propose o novel approach to fine-tune \bert in a structured manner. Specifically, we focus on Large Scale Multilabel Text Classification (\lmtc) where documents are assigned with one or more labels from a large predefined set of hierarchically organized labels. Our approach guides specific \bert layers to predict labels from specific hierarchy levels. Experimenting with two \lmtc datasets we show that this structured fine-tuning approach not only yields better classification results but also leads to better parameter utilization.
\end{abstract}

\section{Introduction}

Despite \bert{'s} \citep{devlin2019} popularity and  effectiveness, little is known about its inner workings. Several attempts have been made to demystify certain aspects of \bert \citep{rogers2020primer}, often leading to contradicting conclusions. For instance, \citet{clark-etal-2019-bert} argue that attention measures the importance of a particular word when computing the next level representation for this word. However, \citet{kovaleva-etal-2019-revealing} showed that most attention heads contain trivial linguistic information and follow a vertical pattern (attention to \cls, \sep, and punctuation tokens), which could be related to under-utilization or over-parameterization issues. Other studies attempted to link specific \bert heads with linguistically interpretable functions \citep{htut2019attention,clark-etal-2019-bert,kovaleva-etal-2019-revealing,voita-etal-2019-analyzing,hoover-etal-2020-exbert,lin-etal-2019-open},  agreeing that no single head densely encodes enough relevant information but instead different linguistic features are learnt by different attention heads. We hypothesize that the aforementioned largely contributes to the lack of attention-based explainability of \bert. Another open topic is how the knowledge is distributed across \bert layers. Most studies agree that syntactic knowledge is gathered in the middle layers \citep{hewitt-manning-2019-structural,goldberg2019assessing,jawahar-etal-2019-bert}, while the final layers are more task-specific. Most importantly, it seems that any semantic knowledge is spread across the model, explaining why non-trivial tasks are better solved at the higher layers \citep{tenney-etal-2019-bert}.

Driven by the above discussion, we propose a novel fine-tuning approach where different parts of \bert are guided to directly solve increasingly challenging classification tasks following an underlying label hierarchy. Specifically, we focus on Large Scale Multilabel Text Classification (\lmtc) where documents are assigned with one or more labels from a large predefined set. The labels are organized in a hierarchy from general to specific concepts. Our approach attempts to tie specific \bert layers with specific hierarchy levels. In effect, each of these layers is responsible for predicting the labels of the corresponding level. We experiment with two \lmtc datasets (\eurlexdata, \mimiciii) and several variations of structured \bert training. Our contributions are: (a) We propose a novel structured approach to fine-tune \bert where specific layers are tied to specific hierarchy levels; (b) We show that structured training yields better results than the baseline across all levels of the hierarchy, while also leading to better parameter utilization.

\begin{figure*}[t!]
    \centering
    \includegraphics[width=\textwidth]{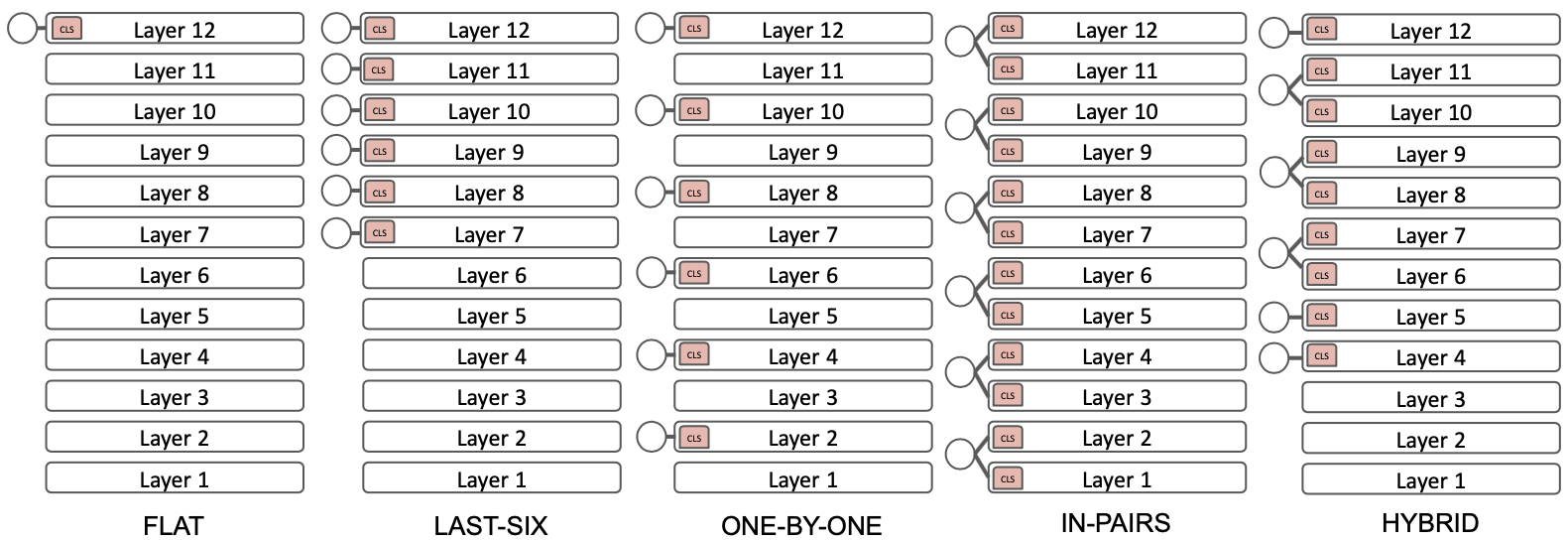}
    \caption{ The five variants of \bert{-based} multi-label classifiers including the flat one and the four structured editions. The circles represent the classification layers attached to \cls tokens across layers.} 
    \label{fig:models}
\end{figure*}

\section{Datasets}
\label{sec:datasets}

\noindent\textbf{\eurlexdata} \cite{chalkidis2019} contains 57k \eu legal acts from \eurlex.\footnote{\url{http://eur-lex.europa.eu/}}
Each act is approx.\ 700 words long and is annotated with one or more concepts from \eurovoc\footnote{\url{http://eurovoc.europa.eu/}} which contains 7,391 concepts organized in an 8-level hierarchy. We truncate the hierarchy to 6 levels by discarding the last 2 levels which contain 50 rarely used labels.\footnote{\label{foot:data_manip}For more details on data manipulation see Appendix A.}\vspace{1mm}

\begin{figure}[t!]
    \centering
    \includegraphics[width=\columnwidth]{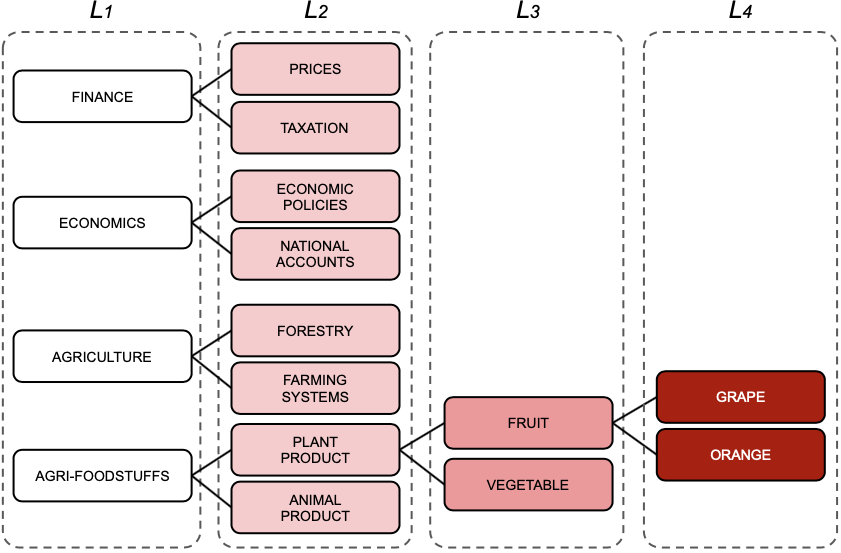}
    \caption{Examples from \eurovoc label hierarchy. Layer-wise models consider all labels in the same level ($L_i$; dashed boxes) of the hierarchy on-par.}
    \label{fig:eurovoc}
\end{figure}

\noindent\textbf{\mimiciii} \cite{Johnson2017} contains approx.\ 52k discharge summaries from \textsc{us} hospitals. Each summary is approx.\ 1.6k words long and is annotated with one or more \icdix\footnote{\url{www.who.int/classifications/icd/en/}} codes. \icdix contains 22,395 codes organized in a 7-level hierarchy. We truncate the hierarchy to 6 levels, discarding the first level which contains only 4 general codes.\textsuperscript{\ref{foot:data_manip}}\vspace{1mm}

\noindent\textbf{Label Augmentation:} In both datasets, we make the assumption that if a label $l$ is assigned to a document then all of its ancestors should also be assigned to this document. Hence, we augment labels by annotating a document with all the ancestors of its assigned labels. For instance, in \eurovoc, if a document is annotated with the label \textit{grape} it will also be annotated with \textit{grape}'s ancestors, i.e., \emph{fruit}, \emph{plant product}, and \emph{agri-foodstuffs} (Figure~\ref{fig:eurovoc}). This assumption is perfectly valid, while also having the added side effect of providing a more accurate test-bed for evaluation. For example, if a classifier mistakenly annotated the document with \textit{citrus fruit}, a sibling of \textit{grape}, in the non-augmented case it would receive a score of zero. By contrast, in the augmented case, assuming it correctly identified all the ancestors of \textit{citrus fruit} it would receive a much higher score of 0.75 having correctly assigned the three ancestors of \textit{grape} but not the (more specialized) label itself. Thus, we believe the model is evaluated more fairly in the augmented case with respect to the hierarchy. This type of evaluation is also in-line with the literature on hierarchical classification \cite{kosmopoulos2015evaluation}.\vspace{1mm}

\begin{table*}[ht!]
    \centering
    \resizebox{\textwidth}{!}{
    \begin{tabular}{l|c|c|c|c|c|c|c|c}
          Labels Depth &  1 & 2 & 3 & 4 & 5 & 6 & Micro & Macro \\
         \hline
         \hline
         \hline
         \multicolumn{9}{c}{\eurlex - \bertbase} \\
         \hline
         \hline
        \#Labels & 21 & 127 & 567 & 3,861 & 2,284 &  481 & 7,341  & 7,341 \\
        \hline
         \baseline & 90.3 $\pm$ 0.2 & 83.9 $\pm$ 0.3 & 81.0 $\pm$ 0.5 & 74.8 $\pm$ 1.0 & 74.5 $\pm$ 1.2 & 79.9 $\pm$ 1.4 & 80.6 $\pm$ 0.6 & 80.7 $\pm$ 0.6\\
         \hline
         \lastsix & \cellcolor{green!50}\textbf{90.7} $\pm$ 0.1 & \cellcolor{green!50}\textbf{84.6} $\pm$ 0.0 & \cellcolor{green!50}\textbf{81.9} $\pm$ 0.2 & \cellcolor{green!50}\textbf{76.8} $\pm$ 0.2 & \cellcolor{green!50}\textbf{77.2} $\pm$ 0.5 & \cellcolor{green!50}\textbf{82.2} $\pm$ 1.1 & \cellcolor{green!50}\textbf{81.7} $\pm$ 0.1 & \cellcolor{green!50}\textbf{82.2} $\pm$ 0.1 \\
         \onebyone & \cellcolor{green!20}90.0 $\pm$ 0.0 & \cellcolor{green!20}84.3 $\pm$ 0.1 & \cellcolor{green!20}81.7 $\pm$ 0.2 & \cellcolor{green!20}76.2 $\pm$ 0.2 & \cellcolor{green!20}76.7 $\pm$ 0.5 & \cellcolor{green!20}81.6 $\pm$ 0.2 & \cellcolor{green!20}81.3 $\pm$ 0.1 & \cellcolor{green!20}81.7 $\pm$ 0.0\\
         \allinpairs  & \cellcolor{green!20}89.9 $\pm$ 0.2 &  \cellcolor{green!20}84.3 $\pm$ 0.2 &  \cellcolor{green!20}81.7 $\pm$ 0.2 & \cellcolor{green!20}76.7 $\pm$ 0.3 & \cellcolor{green!50}\textbf{77.2} $\pm$ 0.6 & \cellcolor{green!20}81.7 $\pm$ 0.4 & \cellcolor{green!20}81.4 $\pm$ 0.1  & \cellcolor{green!20}81.9 $\pm$ 0.4\\
         \skipboost & \cellcolor{green!20}90.5 $\pm$ 0.2 & \cellcolor{green!20}84.3 $\pm$ 0.1 & \cellcolor{green!20}81.7 $\pm$ 0.2 & \cellcolor{green!20}76.6 $\pm$ 0.4 & \cellcolor{green!20}76.6 $\pm$ 0.7 &  \cellcolor{green!20}81.8 $\pm$ 0.7	 & \cellcolor{green!20}81.5 $\pm$ 0.2 & \cellcolor{green!20}81.9 $\pm$ 0.0 \\
         \hline
         \hline
         \multicolumn{9}{c}{\mimiciii - \scibert} \\
         \hline
         \hline
         \#Labels & 79 & 589 & 3,982 & 9,640 & 7,234 &  867 &  22,391 &  22,391 \\
        \hline
         \baseline & 75.5 $\pm$ 0.0 & 66.5 $\pm$ 0.2 & 57.7 $\pm$ 0.4 & 50.8 $\pm$ 0.4 & 43.2 $\pm$ 0.8 & 38.6 $\pm$ 3.3 & 60.1 $\pm$ 0.4 & 55.4 $\pm$ 0.9 \\
         \hline
         \lastsix & \cellcolor{green!50}\textbf{76.8} $\pm$ 0.2 & \cellcolor{green!50}\textbf{67.3} $\pm$ 0.1 & \cellcolor{green!50}\textbf{58.5} $\pm$ 0.1 & \cellcolor{green!20}51.2 $\pm$ 0.0 & \cellcolor{green!50}\textbf{43.8} $\pm$ 0.3 & \cellcolor{green!20}40.9 $\pm$ 0.1 & \cellcolor{green!50}\textbf{60.4} $\pm$ 0.0 & \cellcolor{green!50}\textbf{56.4} $\pm$  0.1 \\
         \onebyone & \cellcolor{green!20}75.7 $\pm$ 0.0 & \cellcolor{green!20}66.7 $\pm$ 0.1  & \cellcolor{green!20}57.9 $\pm$ 0.1 & \cellcolor{red!20}50.6 $\pm$ 0.1 & \cellcolor{green!20}43.4 $\pm$ 0.3 & \cellcolor{green!50}\textbf{41.8} $\pm$ 0.6 & \cellcolor{red!20}59.8 $\pm$ 0.1 & \cellcolor{green!20}56.0 $\pm$ 0.2\\
         \allinpairs & \cellcolor{green!20}75.6 $\pm$ 0.1 &  66.5 $\pm$ 0.1 & \cellcolor{green!20}58.1 $\pm$ 0.1 & \cellcolor{green!20}50.9 $\pm$ 0.1 &  \cellcolor{green!20}43.6 $\pm$ 0.3 & \cellcolor{green!20}40.9 $\pm$ 0.8 & \cellcolor{red!20}59.9 $\pm$ 0.1 & \cellcolor{green!20}55.9 $\pm$  0.2\\
         \skipboost & \cellcolor{green!20}76.4 $\pm$ 0.1 & \cellcolor{green!20}67.0 $\pm$ 0.1 & \cellcolor{green!50}\textbf{58.5} $\pm$ 0.1 & \cellcolor{green!50}\textbf{51.3} $\pm$ 0.0 & \cellcolor{green!50}\textbf{43.8} $\pm$ 0.2	& \cellcolor{green!20}40.0 $\pm$ 0.4 & \cellcolor{green!20}60.3 $\pm$ 0.0 & \cellcolor{green!20}56.2 $\pm$ 0.1 \\ 
    \end{tabular}
    }
    \caption{$\mathrm{R}$-$\mathrm{Precision}$ $\pm$ std for all variants of \bertbase on \eurlex and \mimiciii test data across hierarchy depths.}
    \label{tab:results}
\end{table*}

\section{Structured Learning with BERT}

Before we proceed with the description of our methods (Figure~\ref{fig:models}), we introduce some notation. Given a label hierarchy $L$ of depth $d$, $L_n$ denotes the set of labels in the $n^\mathrm{th}$ level of this hierarchy ($n \leqslant d$). Also, $f_i = \sigma(W_i\cdot c_i + b_i)$ is a classification function, where $W_i$ and $b_i$ are trainable parameters, $c_i$ is the \cls token in the $i^\mathrm{th}$ \bert layer,\footnote{We use \bertbase (12 layers, 768 units, 12 heads).} and $\sigma$ is the sigmoid activation function. Note that the sizes of $W_i$ and $b$ depend on the number of labels that $f_i$ is responsible for predicting, i.e., if $f_i$ predicts the labels of $L_n$, $W_i\in\mathbb{R}^{|L_n|\times768}$ and $b_i\in\mathbb{R}^{|L_n|\times 1}$.\vspace{1mm}

\noindent\textbf{\baseline:} This is a simple baseline which uses $f_{12}$ to predict all labels in the hierarchy in a flat manner. In effect, $W_{12}\in\mathbb{R}^{|L|\times768}$ and $b_i\in\mathbb{R}^{|L|\times 1}$. Note that this model achieves state-of-the-art results on \eurlexdata but not on \mimiciii \cite{chalkidis2020empirical}. However, our results are not directly comparable to \newcite{chalkidis2020empirical} because our methods operate on augmented label sets.\vspace{1mm}

\noindent\textbf{\lastsix:} This method uses the classifiers $f_7$ through $f_{12}$ to predict the labels in $L_1$ through $L_6$, respectively. Our intuition is that the layers 1-6 will retain and enhance their pre-trained functionality, i.e., syntactic knowledge, contextualized representations, while layers 7-12 will leverage this knowledge to better solve their individual tasks. We also expect that the model will show higher parameter utilization for the layers 7-12 since they are forced to solve gradually more refined classification tasks.\vspace{1mm}

\noindent\textbf{\onebyone:} This method utilizes the full depth of \bert in a \emph{``skip one, use one''} fashion, i.e., it uses classifiers $f_i$, $i \in \{2, 4, \dots, 12\}$. In effect, the odd layers ($1, 3, \dots, 11$) are updated only indirectly, through the classification tasks of the even layers. We expect that the odd layers will learn rich latent representations to facilitate the classifiers of the even layers. Spreading the classification tasks across the whole depth of the model will potentially lead to better parameter utilization. On the other hand, it could harm the model's pre-trained functionality and hence its performance.\vspace{1mm}

\noindent\textbf{\allinpairs}: This method also exploits the full depth of \bert, but now the layers are grouped in 6 pairs, $p_n \in \{(1,2), (3,4), \dots, (11,12)\}$. The classifier responsible for the labels of $L_n$ operates on the concatenated \cls tokens of the corresponding pair, e.g., $f_1=\sigma(W_1\cdot [c_1;c_2] + b_1)$ is trained on the labels of $L_1$. We expect \allinpairs to have better parameter utilization than \onebyone, although the risk of hindering performance is now even higher.\vspace{1mm}

\noindent\textbf{\skipboost:} Similarly to \lastsix, this method skips some of the lower \bert layers (3 instead of 6). Also, it ties $L_1$, $L_2$, and $L_6$, which are the hierarchy levels with the fewest labels to layers 4, 5, and 12, respectively. Finally, similarly to \allinpairs the remaining \bert layers are grouped in pairs and are tied to the rest of the hierarchy levels. We expect the first three layers to retain and enhance their pre-trained functionality, while the hierarchy levels with a large number of labels will benefit from the additional parameters at their disposal. 

\begin{figure*}[ht!]
    \centering
    \includegraphics[width=\textwidth]{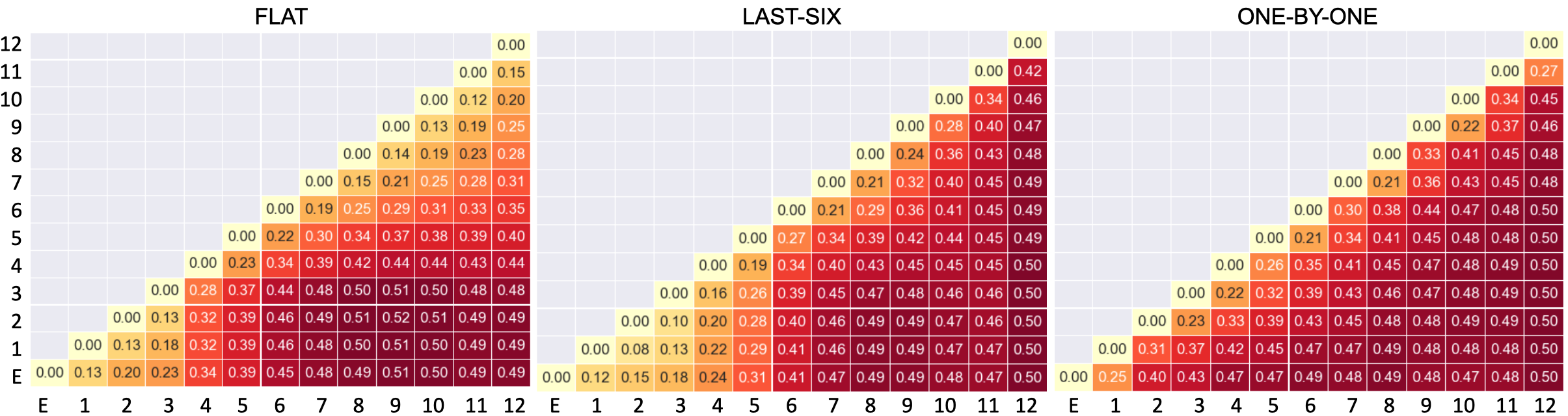}
    \caption{Angular distance between \cls representations across layers.}
    \label{fig:cls_diff}
\end{figure*}

\begin{figure*}[ht!]
    \centering
    \includegraphics[width=\textwidth]{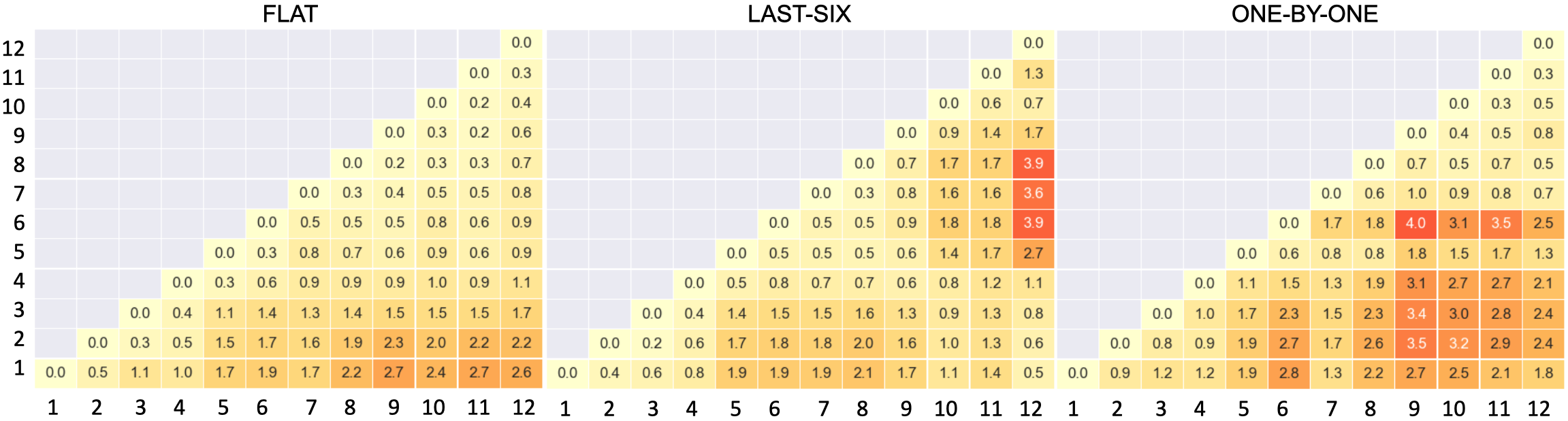}
    \caption{$\mathrm{KL}$-$\mathrm{Divergence}$ between attention distributions across layers.}
    \label{fig:kl}
\end{figure*}

\begin{table*}[ht!]
    \centering
    \resizebox{\textwidth}{!}{
    \begin{tabular}{l|c|c!{\vrule width 2pt}c|c!{\vrule width 2pt}c|c!{\vrule width 2pt}c|c!{\vrule width 2pt}c|c!{\vrule width 2pt}c|c!{\vrule width 2pt}!{\vrule width 2pt}c|c!{\vrule width 2pt}c|c!{\vrule width 2pt}c|c!{\vrule width 2pt}c|c!{\vrule width 2pt}c|c!{\vrule width 2pt}c|c}
          Layer & \multicolumn{2}{c!{\vrule width 2pt}}{1} & \multicolumn{2}{c!{\vrule width 2pt}}{2} & \multicolumn{2}{c!{\vrule width 2pt}}{3} & \multicolumn{2}{c!{\vrule width 2pt}}{4} & \multicolumn{2}{c!{\vrule width 2pt}}{5} & \multicolumn{2}{c!{\vrule width 2pt}!{\vrule width 2pt}}{6} & \multicolumn{2}{c!{\vrule width 2pt}}{7} & \multicolumn{2}{c!{\vrule width 2pt}}{8} & \multicolumn{2}{c!{\vrule width 2pt}}{9} & \multicolumn{2}{c!{\vrule width 2pt}}{10} & \multicolumn{2}{c!{\vrule width 2pt}}{11} & \multicolumn{2}{c}{12} \\
          \hline

\baseline & 4.7 & 1.3 & 3.9 & 0.7 & 2.5 & 0.9 & 3.6 & 1.7 & 3.3 & 2.6 & 3.2 & 3.2 & 3.4 & 3.1 & 2.7 & 2.6 & 2.1 & 2.1 & 2.6 & 1.8 & 2.2 & 1.2 & 2.4 & 0.6 \\
\hline
\lastsix & 4.6 & \cellcolor{yellow!20}\textbf{1.5} & \cellcolor{yellow!50}\textbf{4.2} & 0.6 & \cellcolor{red!50}\textbf{4.1} & 0.8 & \cellcolor{yellow!50}\textbf{3.9} & 1.1 & 2.7 & \cellcolor{yellow!20}\textbf{2.7} & 2.9 & \cellcolor{yellow!20}\textbf{3.6} & 2.9 & 3.0 & 2.6 & \cellcolor{yellow!20}\textbf{2.9} & \cellcolor{orange!50}\textbf{3.0} & \cellcolor{orange!50}\textbf{4.2} & \cellcolor{red!50}\textbf{4.5} & \cellcolor{red!50}\textbf{3.9} & \cellcolor{red!50}\textbf{4.3} & \cellcolor{red!50}\textbf{2.3} & \cellcolor{red!50}\textbf{5.2} & 0.6 \\
\onebyone & 4.5 & \cellcolor{yellow!20}\textbf{1.9} & \cellcolor{yellow!20}\textbf{4.1} & \cellcolor{yellow!20}\textbf{2.4} & \cellcolor{red!50}\textbf{4.0} & \cellcolor{red!50}\textbf{4.7} & \cellcolor{yellow!50}\textbf{4.0} & \cellcolor{yellow!50}\textbf{3.8} & \cellcolor{yellow!50}\textbf{3.7} & \cellcolor{yellow!50}\textbf{4.2} & 2.8 & 2.9 & \cellcolor{red!50}\textbf{4.6} & 2.7 & \cellcolor{red!50}\textbf{4.0} & \cellcolor{red!50}\textbf{3.0} & \cellcolor{red!50}\textbf{3.9} & 1.7 & \cellcolor{red!50}\textbf{3.9} & 1.8 & \cellcolor{red!50}\textbf{4.4} & \cellcolor{red!50}\textbf{1.4} & \cellcolor{red!50}\textbf{4.5} & 0.2 \\
\allinpairs & 4.4 & \cellcolor{yellow!20}\textbf{1.9} & 3.9 & \cellcolor{yellow!20}\textbf{3.0} & \cellcolor{red!50}\textbf{3.9} & \cellcolor{red!50}\textbf{4.6} & \cellcolor{yellow!50}\textbf{4.1} & \cellcolor{yellow!50}\textbf{4.2} & 3.2 & \cellcolor{yellow!20}\textbf{3.6} & \cellcolor{yellow!50}\textbf{3.5} & \cellcolor{yellow!50}\textbf{3.5} & \cellcolor{orange!20}\textbf{4.1} & 2.5 & \cellcolor{orange!50}\textbf{3.5} & \cellcolor{orange!50}\textbf{2.8} & \cellcolor{red!50}\textbf{3.7} & 2.0 & \cellcolor{red!50}\textbf{3.7} & 1.4 & \cellcolor{red!50}\textbf{4.2} & 1.1 & \cellcolor{red!50}\textbf{5.1} & 0.2 \\
\skipboost & 4.6 & \cellcolor{yellow!20}\textbf{2.0} & \cellcolor{orange!20}\textbf{4.5} & \cellcolor{orange!20}\textbf{0.9} & \cellcolor{red!50}\textbf{4.2} & \cellcolor{red!50}\textbf{1.7} & \cellcolor{orange!20}\textbf{4.2} & \cellcolor{orange!20}\textbf{2.1} & \cellcolor{yellow!50}\textbf{3.8} & \cellcolor{yellow!50}\textbf{3.4} & 3.1 & \cellcolor{yellow!20}\textbf{3.7} & \cellcolor{orange!20}\textbf{4.1} & \cellcolor{orange!20}\textbf{3.8} & \cellcolor{red!50}\textbf{3.8} & \cellcolor{red!50}\textbf{3.6} & \cellcolor{red!50}\textbf{4.1} & \cellcolor{red!50}\textbf{2.8} & \cellcolor{red!50}\textbf{4.3} & \cellcolor{red!50}\textbf{2.5} & \cellcolor{red!50}\textbf{4.1} & 1.1 & \cellcolor{red!50}\textbf{4.7} & 0.3 \\
    \end{tabular}
    }
    \caption{$\mathrm{Entropy}$ (left column) and $\mathrm{KL}$-$\mathrm{Divergence}$ (right column) of attention distributions per layer.}
    \label{tab:attention_distribs}
\end{table*}

\section{Experiments}
We report $\mathrm{R}$-$\mathrm{Precision}$ \cite{Manning2009} at each hierarchy level as well as micro (flat) and macro averages across all levels.\footnote{See Appendices B and C for a detailed description on experimental setup and a discussion on \lmtc evaluation.} Table \ref{tab:results} shows the results in both datasets. In \eurlexdata our structured methods always outperform the baseline mostly by a large margin. \lastsix achieves the best overall results and is superior than the other structured methods in all hierarchy levels indicating that allowing the lower layers to retain and enhance their pre-trained functionality is crucial. Similar observations can be made for \mimiciii, but in this case the importance of not damaging \bert{'s} pre-trained functionality is even higher, as evident by the only minor improvements \onebyone and \allinpairs have compared to \baseline.\footnote{This is probably due to the additional difficulties of the clinical domain. See Appendix D for a discussion.} 
\section{Discussion on model utilization}

\noindent\textbf{Comparing \cls representations:} Figure~\ref{fig:cls_diff} shows the average angular distances between the \cls representations of each layer on development data of \eurlexdata.\footnote{\label{foot:cls_heat}For brevity, we only show the heatmaps of three methods. We include the missing ones in Appendix E.} The angular distance is calculated on unit ($\mathrm{L}_2$ normalized) vectors, takes values in $[0, 1]$, and a distance of 0.5 indicates an angle of $90^{\circ}$. We observe that \onebyone leads to larger angles between the representations than \lastsix which in turn yields larger angles than \baseline. In effect, \onebyone and to a lesser extent \lastsix lead to a better parameter utilization than \baseline. To better support this claim we provide a geometric interpretation. We first $\mathrm{L}_2$ normalize all \cls representations. Each normalized representation can be interpreted as a vector having its initial point at the origin and its terminal point at the surface of a 768-dimensional hyper-sphere (centered at the origin). The larger the angle between two \cls vectors the further apart they are on the hyper-sphere's surface. Effectively, \cls vectors with large angles between them cover a larger sub-area of the hyper-sphere's surface indicating that the vector space is utilized to a higher extent which directly implies better parameter utilization.\vspace{1mm}

\noindent\textbf{Comparing attention distributions:} Figure~\ref{fig:kl} shows the $\mathrm{KL}$-$\mathrm{Divergence}$ of the average (across heads) attention for all layers on the development data of \eurlexdata.\textsuperscript{\ref{foot:cls_heat}} A high $\mathrm{KL}$-$\mathrm{Divergence}$ indicates that two layers attend to different sub-word units. Moreover, Table~\ref{tab:attention_distribs} reports the entropy (left column per layer) of the average (across heads) attention distribution per layer. A high entropy indicates that a layer attends to more sub-word units. Table~\ref{tab:attention_distribs} also reports the average $\mathrm{KL}$-$\mathrm{Divergence}$ (right column per layer) between the attention distributions of each possible pair of heads in a layer. A high $\mathrm{KL}$-$\mathrm{Divergence}$ indicates that each head attends to different sub-word units. A first observation is that \baseline attends to almost the same sub-word units across layers (small entropy differences and $\mathrm{KL}$-$\mathrm{Divergence}$ across layers). Interestingly, the different attention heads focus on different sub-word units only in the middle layers (5-8). On the other hand, all the structured methods show better utilization of the attention mechanism, having higher entropy and $\mathrm{KL}$-$\mathrm{Divergence}$ both across heads (Table~\ref{tab:attention_distribs}) and across layers (Figure~\ref{fig:kl}).

\section{Related Work}
Our approach is similar to \citet{wehrmann18a} but they experiment with fully connected networks, which are not well suited for text classification, contrary to stacked transformers \citep{Vaswani2017,devlin2019}. Similarly, \citet{Yan2015} used Convolutional Neural Networks, albeit with shallow hierarchies (2 levels). Although our approach leverages the label hierarchy it should not be confused with hierarchical classification methods \citep{silla2011survey}, which typically employ one classifier per node and cannot scale-up to large hierarchies when considering neural classifiers. A notable exception is the work of \citet{You2019} who employed one bidirectional \lstm with label-wise attention \citep{you2018attentionxml} per hierarchy node. However, for their method to scale-up, they use probabilistic label trees \citep{Khandagale2019} to organize the labels in their own shallow hierarchy which does not follow the abstraction level of the original hierarchy. To the best of our knowledge we are the first to apply this approach to pre-trained language models.

\section{Conclusions and Further Work}
We proposed a novel guided approach to fine-tune \bert, where specific layers are tied to specific hierarchy levels. Experimenting with two \lmtc datasets, we showed that structured training not only yields better results than a flat baseline, but also leads to better parameter utilization. In the future we will try to further increase the parameter utilization by guiding \bert{'s} attention heads to explicitly focus on specific hierarchy parts. We also plan to improve the explainability of our methods with respect to the utilization of their parameters.

\bibliographystyle{acl_natbib}
\bibliography{emnlp2020}

\appendix

\section{Data manipulation}
\label{sec:data_man}
\noindent\textbf{Hierarchy truncation:} In order to directly apply all our methods, we truncate both hierarchies and reduce their depth to six. We believe this truncation is justified since in \eurovoc the last two layers contain a very small number of labels, which are rarely, if at all, assigned and in \icdix the first layer also contains a very small number of labels which are very general and can be trivially classified (Table~\ref{tab:hierarchies}). In both cases it seems that only minimal information is lost which would have small practical use in the classification tasks.\vspace{1mm}

\begin{table}[ht]
    \centering
    \small{
    \begin{tabular}{c|c|c}
         Depth & \eurovoc & \icdix \\
         \hline
         1 & 21  & 4* \\
         2 & 127 & 79 \\
         3 & 568 & 589 \\
         4 & 4,545 & 3,982 \\
         5 & 2,335 & 9,640 \\
         6 & 497 & 7,234 \\
         7 & 79* & 867 \\
         8 & 6* & - \\
         \hline
         Overall & 8,178 / 8,093 & 22,395 / 22,391\\
    \end{tabular}
    }
    \caption{Label distribution across \eurovoc and \icdix hierarchy levels. Concepts (labels) are arranged from more abstract (level 1-2) to more specialized ones (levels 6-8). Labels with an asterisk are truncated in our experiments.}
        \vspace{-2mm}
    \label{tab:hierarchies}
\end{table}

\noindent\textbf{Document Truncation:} Documents in both datasets are often above the 512 token limit of \bert. To reduce document size, we perform a number of pre-processing normalizations, including removal of numeric tokens, punctuation and stop-words.\footnote{Similar procedures are very common in classification, thus we believe they do not harm text semantics.} In \eurlex documents have been tokenized using SpaCy's default tokenizer,\footnote{\url{https://spacy.io}} while in \mimiciii, we use regular expressions tailored for the biomedical domain. While document length is severely reduced post normalization, if a document still has a larger number of tokens, i.e. more than 512, we use the first 512 tokens and ignore the rest.

\section{Experimental Setup}
All our methods build on \bertbase and are implemented in Tensorflow 2. For \eurlex we use the original \bertbase \citep{devlin2019}, while for \mimiciii we use \scibert \citep{beltagy2019}, which has the same architecture (12 layers, 768 hidden units, 12 attention heads), and better suits biomedical documents.\footnote{We use the Transformers library of Huggingface (\url{https://github.com/huggingface/transformers}).} Our models are tuned by grid searching three learning rates ($2\mathrm{e}\text{-}5, 3\mathrm{e}\text{-}5, 5\mathrm{e}\text{-}5$) and two drop-out rates ($0, 0.1$). We use the Adam optimizer \cite{Kingma2015} with early stopping on validation loss. In preliminary experiments, we found that weighting individual losses with respect to the number of labels in each level is crucial. We therefore weigh each loss by the percentage of labels at the corresponding level, i.e.,  $ w_n = \frac{|L_n|}{|L|}$,  where $|L_n|$ is the number of labels in the $n^\mathrm{th}$ level of the hierarchy and $|L|$ is the total number of labels across all levels, e.g., in \eurlexdata, $w_1 = \frac{21}{8093} \approx 0.0026$.

\section{Evaluation in LMTC}
\label{sec:evaluation}
The literature of \lmtc \cite{Rios2018-2, chalkidis2019} mostly uses information retrieval evaluation measures. We support the premise that when the number of labels is that large the problem mimics retrieval with each document acting as a query and the model having to score relevant labels higher than the rest. However in our study, it would be really confusing to report the standard retrieval metrics $\mathrm{Recall}@R$, $\mathrm{Precision}@K$, $\mathrm{nDCG}@K$ since we evaluate our classifiers at each hierarchy depth and reasonable values for $K$ have large fluctuations between levels, as the number of labels per level vastly varies (see Table~\ref{tab:hierarchies}). Instead, we prefer $\mathrm{R}$-$\mathrm{Precision}$ \cite{Manning2009}, which is the $\mathrm{Precision}@R$ where $R$ is the number of gold labels associated with each document. It follows that $\mathrm{R}$-$\mathrm{Precision}$ can neither under-estimate (penalize) nor over-estimate the performance of the models \cite{chalkidis2019}. 

\section{Peculiarities of \mimiciii dataset}
\label{sec:mimic}
In our experiments we observe a hindered performance in \mimiciii, which can be attributed to a number of characteristics of the dataset. Firstly, documents contain a lot of non-trivial biomedical terminology which naturally makes the classification task more difficult. Further, discharge summaries describe a patient's condition during their hospitalization and therefore proper label annotations change throughout the document as the patient's diagnosis changes or as they exhibit new symptoms, e.g., ``\emph{the patient was admitted to the hospital with \underline{no heart issues}, [\dots] the patient had \underline{a heart failure} and died.}''. Both the in-domain language and the constant change of events make the dataset more challenging than \eurlexdata, where documents are more organized and well-written also with simpler language. 

It therefore seems reasonable that in \mimiciii allowing lower \bert layers to retain and enhance the preliminary functionality, without explicitly guiding them, is of utmost importance. We would like to highlight that even though we use \scibert \citep{beltagy2019}, which is based on a new scientific vocabulary, we observe that specialized biomedical terms are often over-fragmented in multiple sub-word units, e.g. `atelectasis' splits into [`ate', `\#\#lect', `\#\#asis']. Thus,  the initial layers need to decipher these over-fragmented sub-word units and reconstruct the original word semantics. On the contrary, in \eurlexdata, classifying general concepts in the initial layers, even considering only the sub-word unit embeddings is plausible.

\begin{figure*}[ht!]
    \centering
    \includegraphics[width=\textwidth]{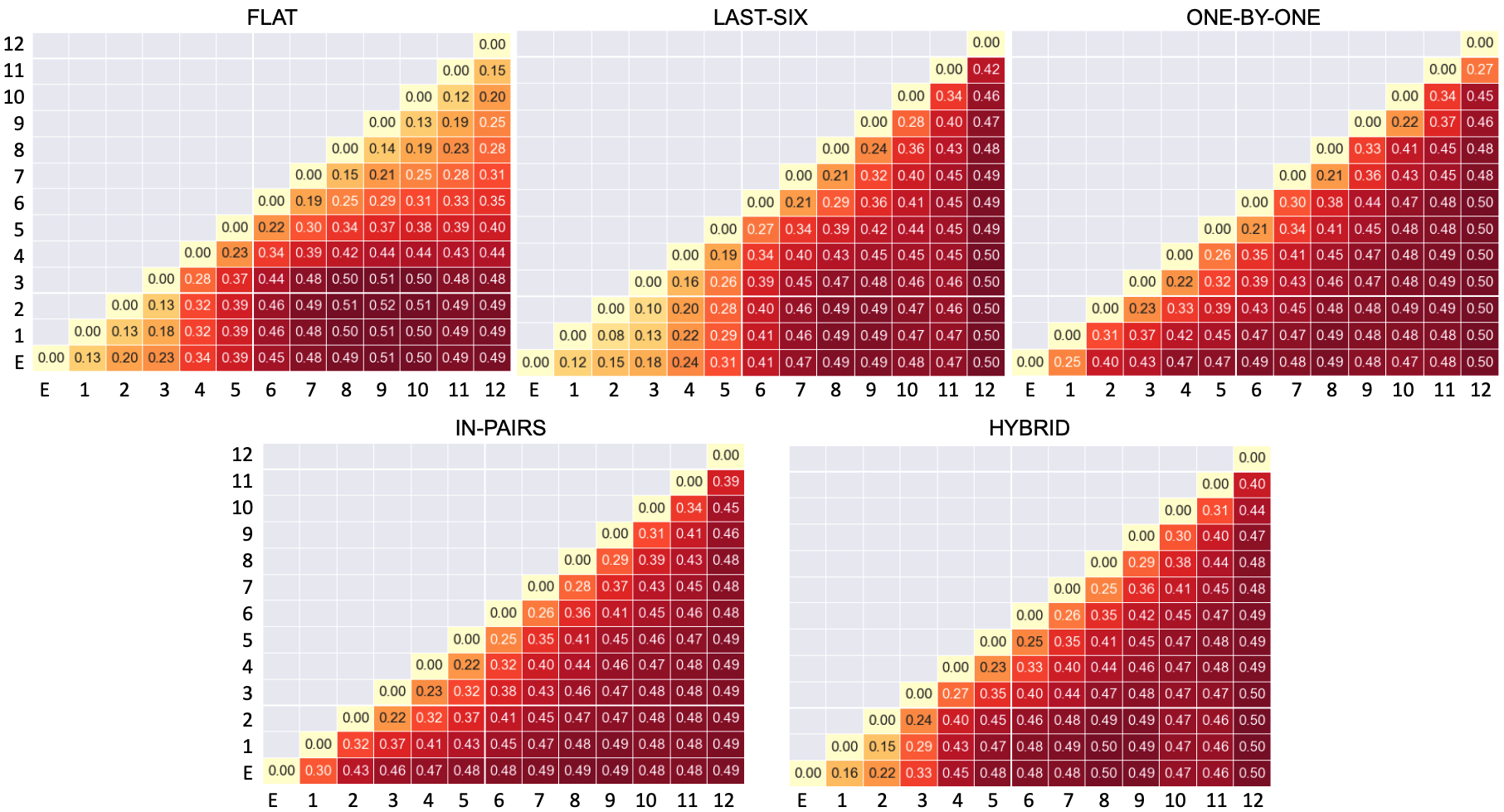}
    \caption{Angular distance between \cls representations across layers in the development dataset of \eurlexdata.}
    \label{fig:cls_app}
\end{figure*}

\begin{figure*}[ht!]
    \centering
    \includegraphics[width=\textwidth]{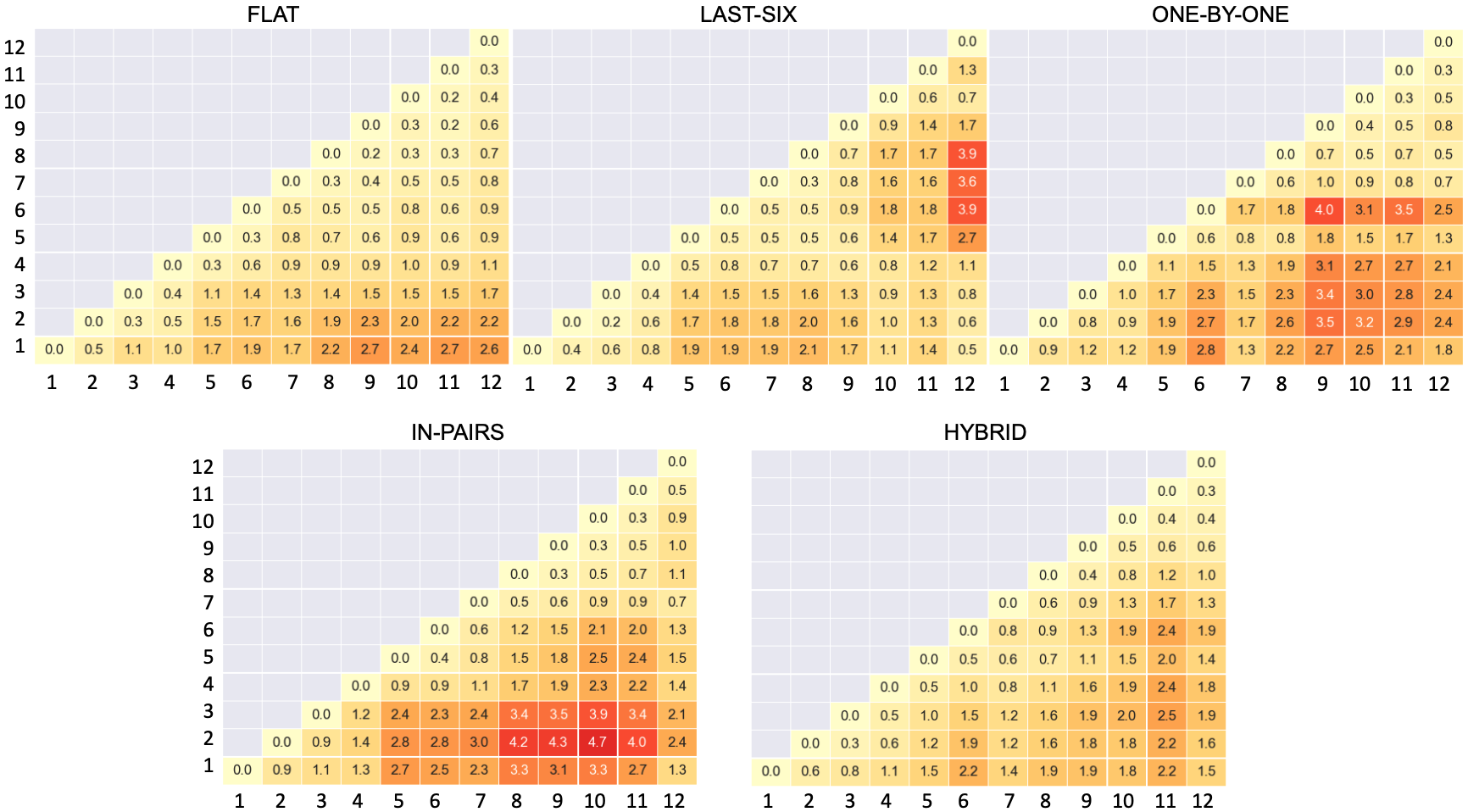}
    \caption{$\mathrm{KL}$-$\mathrm{Divergence}$ between attention distributions across layers in the development dataset of \eurlexdata.}
    \label{fig:kl_app}
\end{figure*}

\begin{figure*}[ht!]
    \centering
    \includegraphics[width=\textwidth]{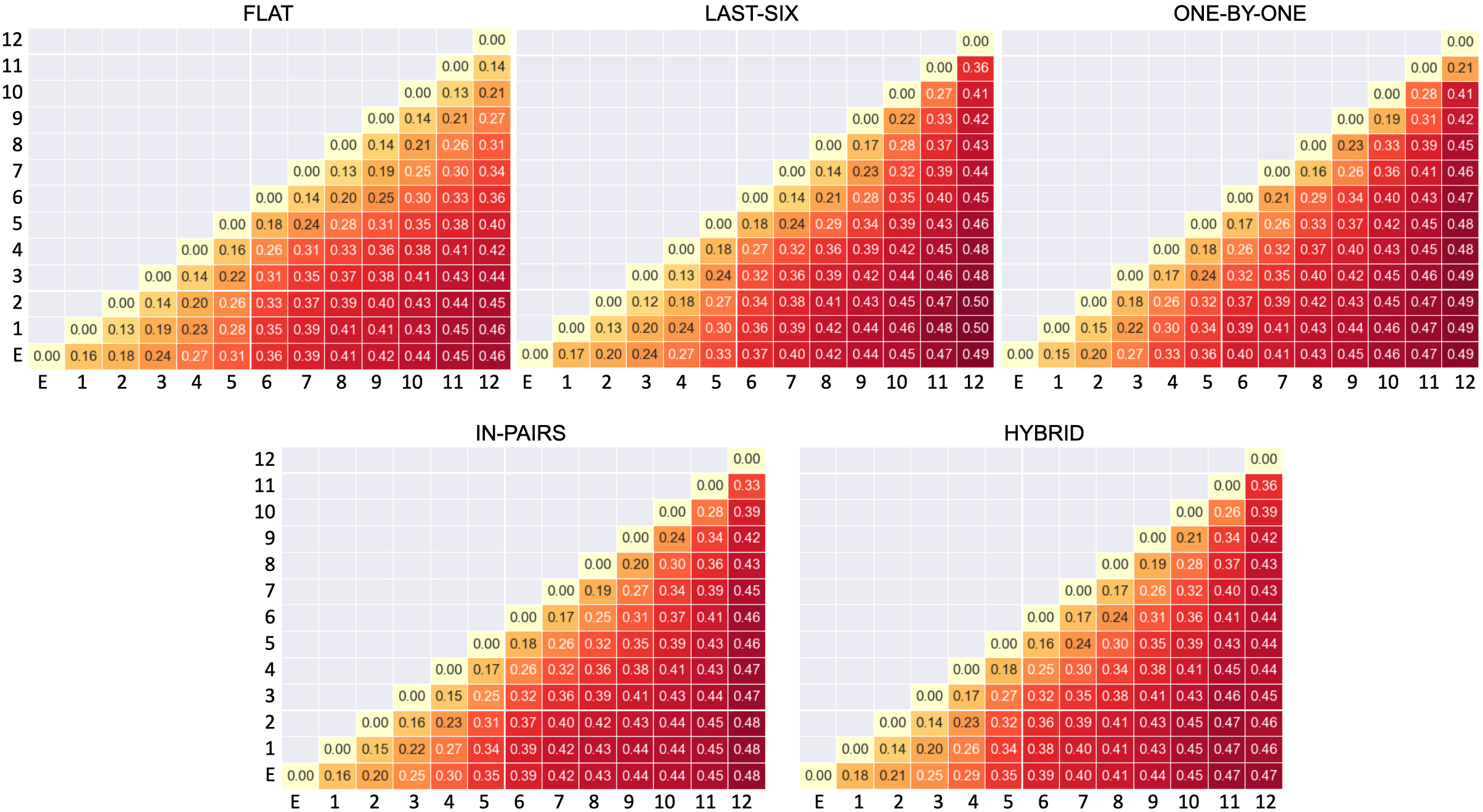}
    \caption{Angular distance between \cls representations across layers in the development dataset of \mimiciii.}
    \label{fig:mimic_cls_app}
\end{figure*}

\begin{figure*}[ht!]
    \centering
    \includegraphics[width=\textwidth]{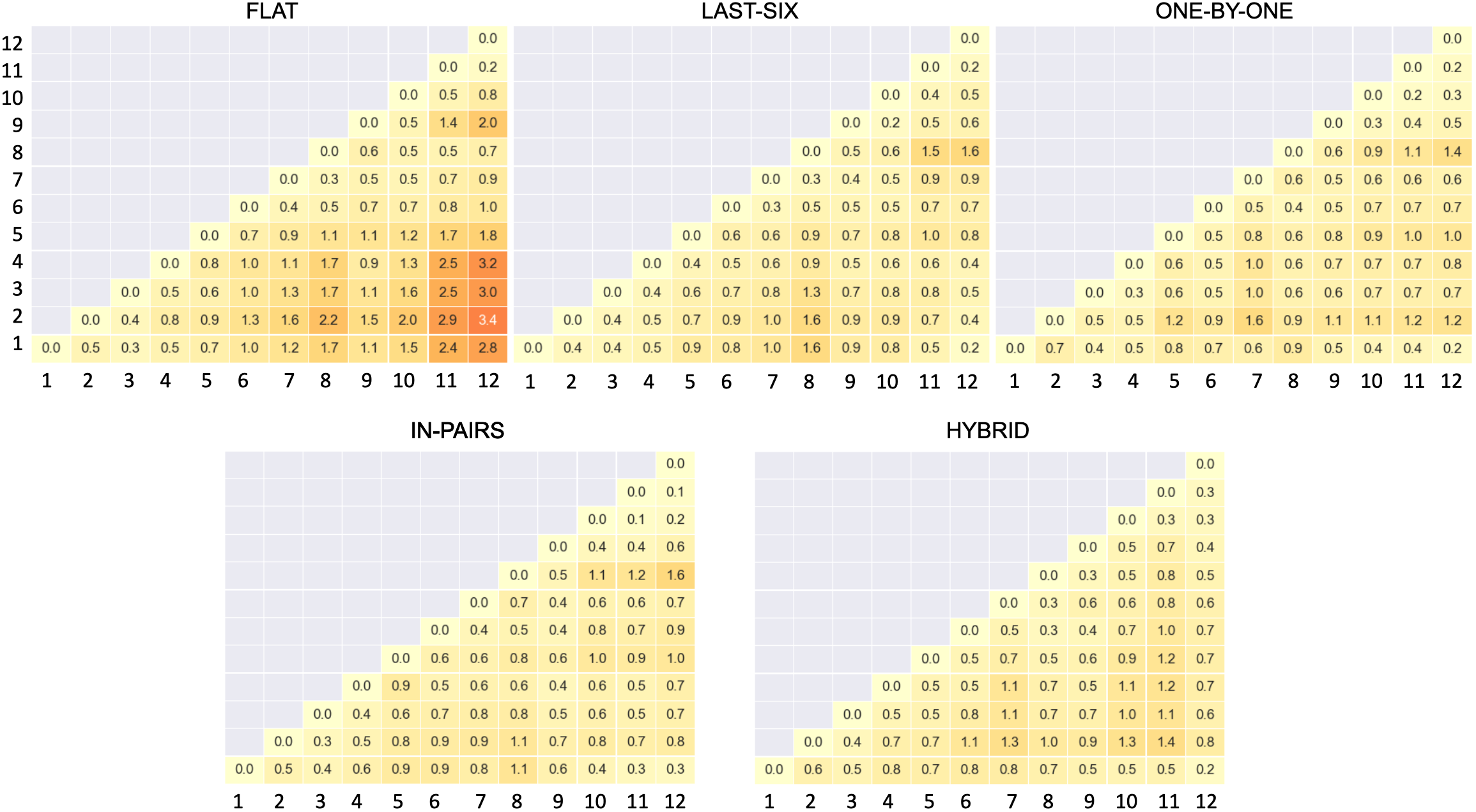}
    \caption{$\mathrm{KL}$-$\mathrm{Divergence}$ between attention distributions across layers in the development dataset of \mimiciii.}
    \label{fig:mimic_kl_app}
\end{figure*}

\section{Discussion on model utilization}

We present additional results for the rest of the methods (\allinpairs, \skipboost). Figure~\ref{fig:cls_app} shows the average angular distances between the \cls representations of each layer (Figure~\ref{fig:cls_app}) for all considered methods.  We observe that the distances of \allinpairs between consecutive \cls representations follow a similar pattern with those of \onebyone, with the exception of 0.25+ distances which are more dense in the upper layers for \allinpairs. This is reasonable, since in \allinpairs all layers directly contribute to the classification tasks. The pattern of \skipboost is very similar to \onebyone and \allinpairs, except for the first three non-guided layers in which distances bear close resemblance to those of the corresponding layers in \lastsix. Similar observations hold for \mimiciii (Figure~\ref{fig:mimic_cls_app}). Finally, Figure~\ref{fig:kl_app} shows the $\mathrm{KL}$-$\mathrm{Divergence}$ of the average (across heads) attention for all layers on the development data. All structured methods show better utilization of the attention mechanism than \baseline, having higher $\mathrm{KL}$-$\mathrm{Divergence}$ across layers. Contrary, in \mimiciii, all structured methods follow a similar pattern of low $\mathrm{KL}$-$\mathrm{Divergence}$ across layers (Figure~\ref{fig:mimic_kl_app}), even lower than the upper layers of \baseline, i.e., the models attend to similar sub-word positions across layers. We aim to further study and explain this behaviour in future work.

\end{document}